%% file: ReLet.tex
\documentclass{article}

\usepackage[preprint]{neurips_2026}

\usepackage{graphicx}
\usepackage[utf8]{inputenc} % allow utf-8 input
\usepackage[T1]{fontenc}    % use 8-bit T1 fonts
\usepackage{hyperref}       % hyperlinks
\usepackage{url}            % simple URL typesetting
\usepackage{booktabs}       % professional-quality tables
\usepackage{amsfonts}       % blackboard math symbols
\usepackage{nicefrac}       % compact symbols for 1/2, etc.
\usepackage{microtype}      % microtypography
\usepackage{xcolor}         % colors
\usepackage{multirow}
\usepackage{adjustbox}
\usepackage{amsmath}
\usepackage{algorithm}
\usepackage{algorithmic}
\usepackage{subcaption}
\usepackage{wrapfig}
\usepackage{graphicx}
\usepackage[most]{tcolorbox}
\usepackage{makecell}
\usepackage{enumitem}
\usepackage{float} 

\setlist[itemize]{noitemsep,leftmargin=*,topsep=0pt}
\setlist[enumerate]{noitemsep,leftmargin=*,topsep=0pt}

\title{Can Test-Time Reconstruction Close the Loop on Latent Reasoning?}

\author{%
  Xiaopeng Yuan$^{1}$,
  Haibo Jin$^{1}$,
  Ye Yu$^{1}$,
  Peng Kuang$^{1}$,
  Lijun Yu$^{1}$,
  Yushun Dong$^{2}$,
  Haohan Wang$^{1}$\thanks{Corresponding author: haohanw@illinois.edu} \\
  $^{1}$University of Illinois Urbana-Champaign \\
  $^{2}$Florida State University
}

\begin{document}

\maketitle

\begin{abstract}

\input{./Sections/Abstract}
\end{abstract}

\section{Introduction}

\input{./Sections/Introduction}

\section{Related Work}

\input{./Sections/Related_Work}

\section{Preliminary}
\input{./Sections/Preliminary}

\section{Reconstruction-Guided Latent Reasoning at Test Time}
\input{./Sections/Cycle_Consistency}

\section{Experiments}
\input{./Sections/Experiments}

\section{Conclusion}
\input{./Sections/Conclusion}

\newpage
% \section*{References}
\bibliographystyle{plainnat}
\bibliography{reference}

\newpage
\appendix
\input{./Sections/Appendix}

%\newpage
%\input{checklist.tex}

\end{document}

%% file: Sections/Abstract.tex
Recent work moves intermediate reasoning from natural-language traces into latent or cache-level representations to reduce token overhead and avoid a discrete communication bottleneck. However, this shift also removes a key advantage of textual reasoning: intermediate states are no longer inspectable, making it difficult to determine whether a latent state still preserves the constraints of the original query. As a result, latent reasoning typically operates in an open loop, where a latent state is produced and consumed without an input-anchored fidelity check. We propose \textbf{ReLAT} (\textbf{Re}construction-Guided \textbf{L}atent Reasoning \textbf{A}t \textbf{T}est Time), a self-supervised test-time training method that closes this loop using the query itself as the reference. Our key observation is that if a latent state faithfully represents a query, the query should be recoverable from it; if the query cannot be recovered, the latent state has lost task-relevant information. ReLAT operationalizes this principle by constructing a differentiable Question $\rightarrow$ Latent Thought $\rightarrow$ Question cycle and optimizing query reconstruction loss through the latent thought before answer generation. This anchors opaque latent computation to the problem specification it is supposed to represent. Across mathematical reasoning, knowledge QA, and code generation benchmarks on the Qwen family, ReLAT consistently improves over single-model inference, text-based collaboration, open-loop latent collaboration, and alternative test-time training objectives. On Qwen3-8B, ReLAT raises AIME 2024 accuracy from 56.7\% to 73.3\%, a 16.6-point gain over the strongest open-loop latent baseline.

%% file: Sections/Introduction.tex
Recent work has increasingly explored \emph{latent reasoning} as an alternative to conventional text-based reasoning. Traditional reasoning paradigms externalize intermediate computation as natural language, including chain-of-thought rationales, self-refinement traces, and text-based communication among agents~\citep{wei2022chain, madaan2023self, shinn2023reflexion, li2023camel, hong2023metagpt}. While such textual traces are readable and externally inspectable, they incur substantial token overhead and force a discrete bottleneck on intermediate computation. A growing line of work therefore moves intermediate reasoning into latent or cache-level representations, including latent collaboration, direct cache-to-cache communication, thought communication, and latent-space inference~\citep{zou2025latent, fu2025cache, zheng2025thought, li2025seek}, aiming to preserve richer internal representations and enable more efficient continuous computation.

This shift, however, removes the one property that made textual reasoning trustworthy: the ability to check, at any intermediate step, whether the reasoning state still corresponds to the original query. A natural-language trace that drops a constraint of the problem is visibly wrong; an opaque latent state that drops the same constraint looks no different from a faithful one, and is consumed by downstream computation as if it were reliable~\citep{dziri2023faith, saparov2022language, peng2025stepwise}. 
Latent reasoning operates in an open loop: the intermediate state is produced and consumed without any check that it still represents the query it was derived from. To close this loop, we need a way to test whether an opaque latent state still preserves the information needed to specify its original query.

The key observation is that the query itself is the natural reference for this test. The information at risk in the latent state is precisely the set of constraints expressed by this particular query, so the query is what the latent state should be measured against. If a latent state is a faithful intermediate representation of a query, then the query should be recoverable from it; if the query cannot be recovered, the latent state has lost task-relevant information regardless of how confident or internally consistent it appears. Existing approaches do not provide this input-anchored fidelity check. This leaves a gap: how can we provide a self-supervised and differentiable fidelity signal that anchors latent states to the original query?
 %Test-time training methods provide instance-level correction through unsupervised auxiliary objectives, but existing objectives are usually based on confidence, entropy minimization, feedback, or local consistency, rather than whether the latent state preserves the constraints of the original query~\citep{zhang2024come, bansal2025let, li2025learning, hubotter2410efficiently} \hwc{I think it's a distration you mention test-time now, you can save it to the related work or somewhere later. For the part, it seems you are trying to say that "X does not exist; TTT is close but it's not X; so we did X" for a reader knows TTT, it's fine, for a reader who doens't know it, it's just ``why do you suddently mention TTT, it's not X anyway'' For readers who know TTT, they will find it later anyway.}. 
 
We propose \textbf{ReLAT} (\textbf{Re}construction-Guided \textbf{L}atent Reasoning \textbf{A}t \textbf{T}est Time), which operationalizes this principle as a closed-loop adaptation procedure. For each test query, ReLAT constructs a differentiable Question $\rightarrow$ Latent Thought $\rightarrow$ Question cycle and minimizes the reconstruction loss on the latent thought before any answer is generated. This anchors the opaque latent state back to the problem specification it is supposed to represent, turning latent reasoning from an open-loop pipeline into a closed-loop one whose intermediate state is verified against the only ground truth available at test time.

Our contributions are three-fold. 
\begin{itemize}
   \item \textbf{Latent reasoning as a closed-loop problem.} We identify the loss of intermediate fidelity as the core cost of moving reasoning from natural language into latent space. We formulate this as an input-fidelity problem: once an opaque latent state is produced, standard latent reasoning pipelines consume it without checking whether it still preserves the query it was derived from.
   \item \textbf{Reconstruction as the natural fidelity signal.} We argue that the original query is the natural reference for testing latent-state fidelity. If a latent state faithfully represents a query, the query should be recoverable from it; if not, the latent state has lost task-relevant information. We instantiate this idea as \textbf{ReLAT}, a differentiable Question $\rightarrow$ Latent Thought $\rightarrow$ Question loop optimized at test time before answer generation.
   \item \textbf{Systematic evaluation.} We benchmark ReLAT on mathematical reasoning, knowledge-intensive QA, and code generation, comparing against single-model inference, text-level communication, static latent-state communication, and alternative test-time training objectives.
\end{itemize}

%% file: Sections/Related_Work.tex
\textbf{Natural-Language Reasoning and Communication.} Early reasoning and multi-agent systems have primarily relied on natural language as the medium for intermediate reasoning and communication. Chain-of-thought prompting, self-reflection, critic-based refinement, debate, and verbal feedback make intermediate reasoning visible and easy to inspect~\citep{NEURIPS2022_9d560961, gou2024criticlargelanguagemodels}. Methods such as ReAct~\citep{yao2022react}, Reflexion~\citep{shinn2023reflexion}, and Self-Refine~\citep{madaan2023self} show that models can improve their outputs by generating, evaluating, and revising textual reasoning traces. Text-based multi-agent systems extend this paradigm by allowing agents to exchange natural-language messages, critiques, or proposals before producing a final answer~\citep{wu2023autogenenablingnextgenllm, hong2024metagptmetaprogrammingmultiagent}. However, explicit text communication is token-intensive and discrete, since each intermediate state must be decoded into language before being passed to another reasoning step or agent. This textualization process can create a lossy communication bottleneck, where information in internal representations is compressed, omitted, or distorted.

\textbf{Latent Reasoning and Communication.} To bypass this textual bottleneck, recent work explores latent or cache-level representations as a more direct medium for intermediate reasoning and communication~\citep{xu2025softcotsoftchainofthoughtefficient, hao2025traininglargelanguagemodels}. ThoughtComm~\citep{zheng2025thought} uses learned latent prefixes to bridge agent states, while Cache-to-Cache~\citep{fu2025cache} synchronizes KV representations between heterogeneous models. LatentMAS~\citep{zou2025latent} further shows that propagating KV-cache-based thoughts can streamline multi-stage reasoning. These methods reflect a broader shift from textual reasoning traces to latent-space computation. However, most latent methods focus on how to transmit internal states rather than how to verify them. Once a hidden state or cache state is generated, it is typically treated as a reliable intermediate state and passed forward. Because these states are opaque, semantic drift may remain undetected and propagate through the reasoning chain.

\begin{figure}[t]
    \centering  \includegraphics[width=\linewidth]{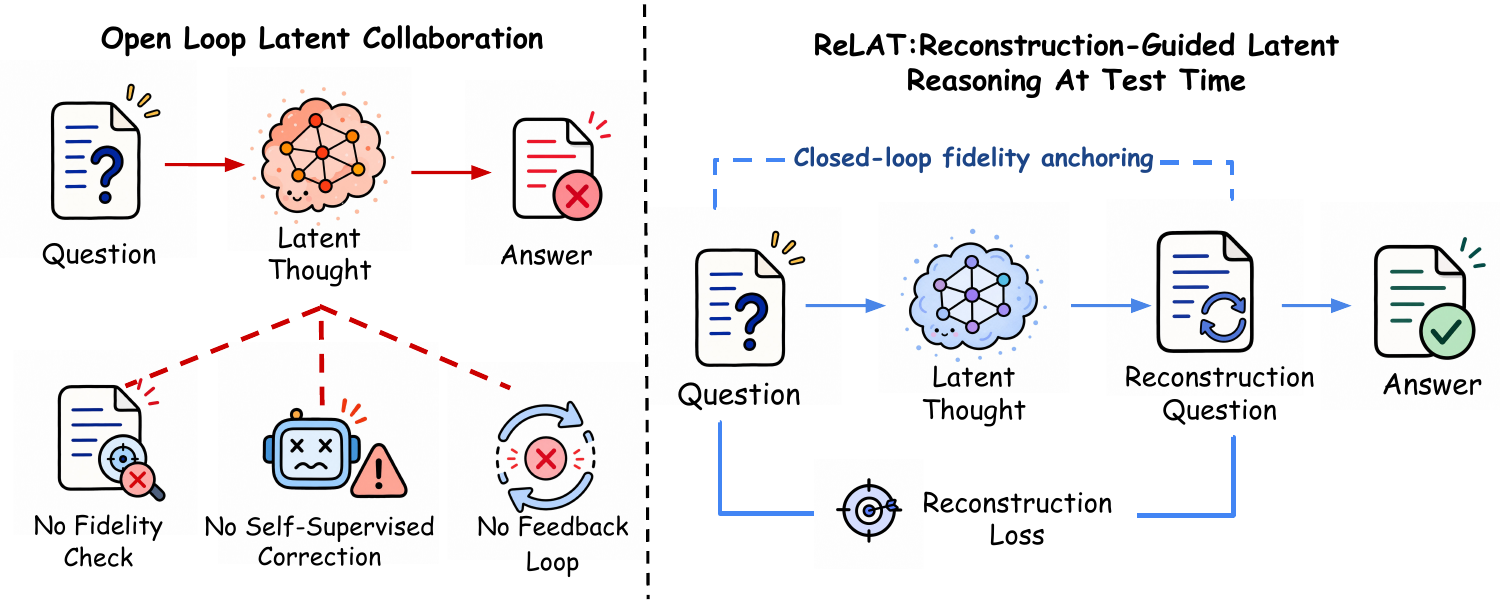}
    \caption{\textbf{Framework Overview.} Comparison between Open-Loop Latent Collaboration and our proposed Reconstruction-Guided Latent Reasoning At Test Time framework.}
    \label{fig:main_framework}
\end{figure}

\textbf{Test-Time Training.} The opacity of latent reasoning raises not only a verification problem, but also an adaptation problem: if a latent state drifts from the original query constraints, the model needs a way to correct this failure during inference without ground-truth answers. Test-time training provides a natural mechanism for such instance-level correction by adapting model behavior to each input using self-supervised or auxiliary objectives. Existing TTT methods often optimize entropy minimization, perplexity minimization, consistency regularization, feedback-based refinement, or query-specific parameter updates. For example, Anti-CF~\citep{su2023beware} and SYTTA~\citep{xu2025you} encourage confident predictions under distribution shift, while COME~\citep{zhang2024come} introduces conservative constraints to reduce collapse. Recent methods such as QTTT~\citep{bansal2025let} study efficient test-time updates through query matrices, and related work explores active data selection or latent optimization for long-context adaptation~\citep{hubotter2024efficiently}. However, these objectives mainly target confidence, consistency, or context adaptation, rather than providing a structural check that latent thoughts preserve the original query constraints.

\textbf{Gap and Motivation.} Together, these works leave a clear gap. Natural-language reasoning is visible but token-intensive. Latent reasoning is efficient but opaque and often open-loop. Test-time training enables instance-specific adaptation, but existing objectives do not directly verify the fidelity of latent reasoning states. ReLAT addresses this gap by introducing a differentiable \emph{Question $\rightarrow$ Latent Thought $\rightarrow$ Question} cycle as a self-supervised fidelity signal. The key intuition is that a useful latent thought should preserve enough information to recover the core constraints of the original question. This reconstruction criterion is not sufficient for correct reasoning, but it provides a necessary structural anchor for filtering and correcting invalid latent states. By optimizing this signal at test time, ReLAT turns open-loop latent reasoning into a closed-loop fidelity-checking process.

%% file: Sections/Preliminary.tex
\subsection{Latent-State Conditioning}
\label{sec:Latent-State}
Latent reasoning methods replace explicit natural-language intermediate traces with continuous intermediate states. Given an input query $Q$, an upstream computation produces a latent thought sequence
$\mathcal{E}=(e_1,e_2,\ldots,e_K)$, where each $e_i\in\mathbb{R}^d$. A downstream computation then conditions on both the original query and the
latent state:
\begin{equation}
    \mathcal{E}=g_\theta(Q), \qquad z=f_\theta([Q;\mathcal{E}]),
\end{equation}
where $[\cdot;\cdot]$ denotes concatenation along the sequence dimension.

We refer to this pattern as \emph{open-loop latent conditioning}. Once $\mathcal{E}$ is produced, the downstream computation consumes it as an intermediate reasoning state without checking whether its content still matches the query that produced it. This is different from ordinary hidden representations used inside a single forward pass: in latent reasoning, $\mathcal{E}$ is explicitly reused as a conditioning signal for later computation, and therefore any missing or distorted constraint can influence the subsequent reasoning trajectory.

The presence of the original query $Q$ in the downstream input does not remove this problem. Because $f_\theta$ conditions on both $Q$ and $\mathcal{E}$, the final output may appear plausible even when the latent state only partially preserves the task specification. In this case, degradation of $\mathcal{E}$ is not directly observable from the latent state itself and may only surface as an downstream reasoning error. We say that $\mathcal{E}$ is \emph{faithful} to $Q$ if it preserves the task-specific constraints expressed in $Q$. Since different queries specify different constraints, latent-state fidelity is inherently query-specific. 

\subsection{Test-Time Training (TTT)}
A direct way to implement such per-query refinement is test-time training, which adapts a model to each test instance through temporary parameter updates without ground-truth labels. Given a test query $Q$, TTT minimizes an unsupervised auxiliary loss $\mathcal{L}_{aux}$ before producing the final answer.

The process consists of two stages:
\begin{enumerate}
    \item \textbf{Optimization:} For a given query $Q$, temporary updates $\Delta \theta$ are computed by minimizing the auxiliary loss:
    \begin{equation}
        \Delta \theta^* \approx \arg\min_{\Delta \theta} 
        \mathcal{L}_{aux}(Q; \theta + \Delta \theta).
    \end{equation}

    \item \textbf{Inference:} The final prediction $y$ is generated using the adapted parameters $\theta' = \theta + \Delta \theta^*$. After inference, the temporary updates are discarded, and the model reverts to the original parameters $\theta$ for the next instance.
\end{enumerate}

In our setting, the temporary updates are restricted to Low-Rank Adapters (LoRA) for efficiency. For a weight matrix $W_0$, the adapted weight is parameterized as
\begin{equation}
    W = W_0 + \Delta W, \quad \Delta W = BA,
\end{equation}
where $A$ and $B$ are low-rank matrices while $W_0$ remains frozen.

TTT itself is a general framework: what property is enforced during adaptation depends on the choice of $\mathcal{L}_{aux}$. Existing TTT objectives such as entropy minimization, confidence-based adaptation, or local consistency target generic prediction stability rather than latent-state fidelity. 

%% file: Sections/Cycle_Consistency.tex
\subsection{Problem Formulation}
Following the open-loop fidelity problem formalized above, ReLAT instantiates the required fidelity signal through input reconstruction. Given a test query $Q$, let $\mathcal{E}_{\phi}(Q)$ denote the latent thought sequence produced under temporary LoRA parameters $\phi$. We say that $\mathcal{E}_{\phi}(Q)$ is faithful to $Q$ if it preserves the query-specific constraints expressed in $Q$.

The faithfulness of $\mathcal{E}_{\phi}(Q)$ is not directly observable at test time, since the constraints encoded in $Q$ are not separately labeled. ReLAT therefore uses $Q$ itself as an observable proxy: because $Q$ specifies the task at hand, the ability to reconstruct $Q$ from $\mathcal{E}_{\phi}(Q)$ provides a self-supervised signal for whether the latent state retains the query-specific constraints needed to support the task. This reconstruction criterion serves as an input-anchored structural check rather than a certificate of correctness. This motivates the reconstruction loop
\begin{equation}
    Q \rightarrow \mathcal{E}_{\phi}(Q) \rightarrow \hat Q,
\end{equation}
which closes the open-loop conditioning of Section~\ref{sec:Latent-State} by routing a feedback signal from $\mathcal{E}_{\phi}(Q)$ back to the original query.

\subsection{Differentiable Thought Representation}
To enable gradient-based optimization during test-time training, we replace discrete token selection with a continuous relaxation. Specifically, we represent the latent thought sequence as $\mathcal{E} = (\hat e_1, \dots, \hat e_K)$, where each $\hat e_j \in \mathbb{R}^d$ is computed as the expected token embedding under the model's vocabulary distribution.

Given the logits $\ell_j \in \mathbb{R}^V$ at latent step $j$, we first compute a softened vocabulary distribution:
\begin{equation}
    \alpha_j = \mathrm{softmax}(\ell_j / \tau),
\end{equation}
where $\tau$ is a temperature hyperparameter. The latent thought vector is then obtained as
\begin{equation}
    \hat e_j = \alpha_j^\top W_{\mathrm{emb}},
\end{equation}
where $W_{\mathrm{emb}} \in \mathbb{R}^{V \times d}$ denotes the model's token embedding matrix. This relaxation keeps each latent vector tied to the model's embedding space while preserving a differentiable path from the ReLAT reconstruction loss to the latent thought generation process.

\subsection{ReLAT: Reconstruction-Guided Latent Reasoning At Test Time}

\begin{wrapfigure}{r}{0.5\linewidth}
\vspace{-20pt}
\begin{minipage}{\linewidth}
\begin{algorithm}[H]
\caption{ReLAT: Reconstruction-Guided Latent Reasoning At Test Time}
\label{alg:relat}
\footnotesize
\begin{algorithmic}[1]
\REQUIRE Query $Q$, model $M_\theta$, LoRA parameters $\phi$, TTT steps $N$, learning rate $\eta$, latent length $K$, temperature $\tau$.
\STATE $\phi_0 \leftarrow \phi$ 
\FOR{$n = 1$ \TO $N$}
    \STATE \textit{// Step 1: Differentiable latent thought generation}
    \STATE $E_Q \leftarrow \text{Embed}(Q)$
    \STATE Initialize $\mathcal{E} = \emptyset, u_0 = E_Q$
    \FOR{$j = 1$ \TO $K$}
        \STATE $\ell_j \leftarrow M_{\theta, \phi_{n-1}}(u_{j-1})$
        \STATE $\alpha_j \leftarrow \text{Softmax}(\ell_j / \tau)$
        \STATE $\hat{\mathbf{e}}_j \leftarrow \alpha_j^{\top} W_{\text{emb}}$
        \STATE $\mathcal{E} \leftarrow \mathcal{E} \oplus \hat{\mathbf{e}}_j, \quad u_j \leftarrow \hat{\mathbf{e}}_j$
    \ENDFOR
    \STATE \textit{// Step 2: Question reconstruction}
    \STATE $\hat{Q} \leftarrow M_{\theta, \phi_{n-1}}([\mathcal{E}])$
    \STATE \textit{// Step 3: LoRA-only test-time update}
    \STATE $\mathcal{L}_{\text{ReLAT}} = \text{MaskedCE}(\hat{Q}, Q)$
    \STATE $\phi_n \leftarrow \text{AdamW}(\phi_{n-1}, \nabla_\phi \mathcal{L}_{\text{ReLAT}}, \eta)$
\ENDFOR
\STATE \textbf{Final Inference:} $y \leftarrow \text{Generate}(M_{\theta, \phi_N}, Q)$
\STATE $\phi \leftarrow \phi_0$
\RETURN $y$
\end{algorithmic}
\end{algorithm}
\end{minipage}
\vspace{-10pt}
\end{wrapfigure}

Building on the differentiable latent representation introduced above, we propose \textbf{ReLAT} (\textbf{Re}construction-Guided \textbf{L}atent Reasoning \textbf{A}t \textbf{T}est Time), a self-supervised test-time training procedure for latent reasoning. ReLAT uses a single backbone model with temporary LoRA adapters to construct a differentiable Question $\rightarrow$ Latent Thought $\rightarrow$ Question loop. The loop is used as an auxiliary adaptation signal before final answer generation.

Given an input question $Q = (q_1, q_2, \ldots, q_m)$, ReLAT first produces a continuous latent thought sequence under the current LoRA parameters:
\begin{equation}
    \mathcal{E}_{\phi}(Q)
    =
    (\hat{\mathbf{e}}_1, \hat{\mathbf{e}}_2, \ldots, \hat{\mathbf{e}}_K),
\end{equation}
where each latent vector $\hat{\mathbf{e}}_j$ is obtained through the soft embedding relaxation defined above. During training, these latent vectors are not decoded into discrete text or detached from the computation graph, so the reconstruction objective can backpropagate through $\mathcal{E}_{\phi}(Q)$ into the LoRA parameters that produced it.

ReLAT then asks the same model to reconstruct the original question from the latent thought sequence. The reconstruction distribution is defined as
\begin{equation}
    P(Q \mid \mathcal{E}; \theta, \phi)
    =
    \prod_{k=1}^{m}
    P(q_k \mid \mathcal{E}, q_{<k}; \theta, \phi),
\end{equation}
where $\theta$ denotes the frozen backbone parameters and $\phi$ denotes the temporary LoRA parameters updated during test-time training. The reconstruction loss is
\begin{equation}
    \mathcal{L}_{\text{ReLAT}}
    =
    - \sum_{k \in \mathcal{M}(Q)}
    \log P(q_k \mid \mathcal{E}, q_{<k}; \theta, \phi),
\end{equation}
where $\mathcal{M}(Q)$ is a loss mask that selects the original question tokens and excludes template tokens, padding, and other formatting artifacts. This objective encourages the latent thought sequence to retain the problem-specific constraints needed to reconstruct the input.

During test-time training, the backbone model remains frozen and only the LoRA parameters are updated:
\begin{equation}
    \phi^{*}
    \approx
    \arg\min_{\phi}
    \mathcal{L}_{\text{ReLAT}}(Q; \theta, \phi).
\end{equation}
After $N$ optimization steps, the final answer is generated from the original question using the adapted model:
\begin{equation}
    y = \mathrm{Generate}(M_{\theta,\phi^{*}}, Q).
\end{equation}
The reconstructed question is used only to define the training loss; the final answer is not generated from $\hat Q$. After inference, the LoRA parameters are reset before processing the next test example, ensuring instance-level training without carrying information across samples.

Thus, ReLAT uses reconstruction as an adaptation signal rather than as a substitute for solving the task. The procedure turns open-loop latent reasoning into a closed-loop test-time process that anchors latent states to the current input specification before final answer generation.

%% file: Sections/Experiments.tex
\subsection{Setup}
\label{sec:experiements}
\textbf{Evaluation Datasets.} We benchmark our approach across three categories: 
(1) \textbf{Mathematical reasoning}, using AIME 2024/2025\citep{mathai_aime2025, maxwelljia_aime2024} for competition-level math problems; 
(2) \textbf{Code generation}, using MBPP+\citep{liu2023your} with augmented test cases to ensure functional correctness; and 
(3) \textbf{Knowledge-based Q\&A}, featuring GPQA-Diamond\citep{rein2024gpqa} for expert-level scientific reasoning and MedQA\citep{jin2021disease} for professional medical knowledge.

\textbf{LLMs and Baselines.} We evaluate ReLAT on four backbone models: Qwen3-4B, Qwen3-8B, Qwen3-14B\citep{yang2025qwen3}, and DeepSeek-R1-Distill-Qwen-7B\citep{deepseekai2025deepseekr1incentivizingreasoningcapability}. These models provide a range of backbone settings with different scales and reasoning capabilities.  
We compare ReLAT with three types of baselines: 
(1) \textbf{Single}, standard single-model inference; 
(2) \textbf{TextMAS}, a text-level communication baseline where intermediate reasoning is exchanged through discrete natural-language messages; and 
(3) \textbf{LatentMAS}\citep{zou2025latent}, an open-loop latent communication baseline where continuous KV-cache states are passed forward and treated as reliable once produced. 
This comparison evaluates whether ReLAT's reconstruction-guided test-time training provides gains beyond standard inference, explicit text-mediated communication, and open-loop latent-state communication.

\textbf{Implementation Details.} We implement ReLAT on all evaluated backbone models, including Qwen3-4B, Qwen3-8B, Qwen3-14B, and DeepSeek-R1-Distill-Qwen-7B. All experiments are conducted on NVIDIA L40S GPUs. During test-time training, the backbone model parameters are frozen and only temporary LoRA parameters are updated. In the differentiable latent loop, the length of the latent thought sequence $\mathcal{E}$ is fixed at $K=16$. We use a Softmax-based continuous relaxation mechanism to maintain gradient flow through latent thought generation.

For the test-time training phase, the learning rate $\eta$ is selected from $[1 \times 10^{-5}, 5 \times 10^{-5}]$, and we use AdamW to perform $N=16$ optimization steps per instance. To accommodate the varying reasoning complexity across benchmarks, we set the maximum generation length to 20,000 tokens for AIME 2024/2025, 8,192 tokens for GPQA-Diamond, and 4,192 tokens for MedQA and MBPP+. The same task prompts and decoding settings are used across methods for a fair comparison. In the final generation stage, greedy decoding is applied to produce the answer. After each test instance, all LoRA parameters are reset to their initial values to preserve independence across examples and prevent cross-instance parameter contamination.

\subsection{Main Results}

\begin{table*}[t]
\centering
\footnotesize
\caption{\textbf{Main Results (Accuracy).} Comparison across Qwen3 series. Improvements are reported relative to the \textbf{Single} baseline. The best results are highlighted in \textbf{bold}.}
\label{tab:main_accuracy_percentage}
\renewcommand{\arraystretch}{1.}
\setlength{\tabcolsep}{5pt} 
\begin{tabular}{ll cccccc}
\toprule
\textbf{Model} & \textbf{Method} & \textbf{AIME 2024} & \textbf{AIME 2025} & \textbf{MedQA} & \textbf{MBPP+} & \textbf{GPQA D.} & \textbf{Avg.} \\
\midrule
\multirow{7}{*}{\makecell[l]{\textbf{Qwen3}\\\textbf{-4B}}}
 & Single & 43.3\% & 43.3\% & 47.7\% & 63.5\% & 36.3\% & 46.8\% \\
 \cmidrule{2-8}
 & \multirow{2}{*}{TextMAS} & 46.7\% & 43.3\% & 65.3\% & 69.8\% & 40.4\% & 53.1\% \\
 & & \scriptsize(+$3.4\%\uparrow$) & \scriptsize(+0.0\%) & \scriptsize(+$17.6\%\uparrow$) & \scriptsize(+$6.3\%\uparrow$) & \scriptsize(+$4.1\%\uparrow$) & \scriptsize(+$6.3\%\uparrow$) \\
 \cmidrule{2-8}
 & \multirow{2}{*}{LatentMAS} & 56.7\% & 50.0\% & 66.3\% & 73.5\% & 41.9\% & 57.7\% \\
 & & \scriptsize(+$13.4\%\uparrow$) & \scriptsize(+$6.7\%\uparrow$) & \scriptsize(+$18.6\%\uparrow$) & \scriptsize(+$10.0\%\uparrow$) & \scriptsize(+$5.6\%\uparrow$) & \scriptsize(+$10.9\%\uparrow$) \\
 \cmidrule{2-8}
 & \multirow{2}{*}{\textbf{Ours}} & \textbf{63.3\%} & \textbf{56.7\%} & \textbf{67.8\%} & \textbf{79.9\%} & \textbf{42.4\%} & \textbf{62.0\%} \\
 & & \textbf{\scriptsize(+$20.0\%\uparrow$)} & \textbf{\scriptsize(+$13.4\%\uparrow$)} & \textbf{\scriptsize(+$20.1\%\uparrow$)} & \textbf{\scriptsize(+$16.4\%\uparrow$)} & \textbf{\scriptsize(+$6.1\%\uparrow$)} & \textbf{\scriptsize(+$15.2\%\uparrow$)} \\
\midrule
\multirow{7}{*}{\makecell[l]{\textbf{Qwen3}\\\textbf{-8B}}}
 & Single & 50.0\% & 46.7\% & 53.0\% & 64.8\% & 39.9\% & 50.9\% \\
 \cmidrule{2-8}
 & \multirow{2}{*}{TextMAS} & 53.3\% & 53.3\% & 75.0\% & 69.5\% & 43.4\% & 58.9\% \\
 & & \scriptsize(+$3.3\%\uparrow$) & \scriptsize(+$6.6\%\uparrow$) & \scriptsize(+$22.0\%\uparrow$) & \scriptsize(+$4.7\%\uparrow$) & \scriptsize(+$3.5\%\uparrow$) & \scriptsize(+$8.0\%\uparrow$) \\
 \cmidrule{2-8}
 & \multirow{2}{*}{LatentMAS} & 56.7\% & 53.3\% & \textbf{75.3\%} & 74.6\% & 45.5\% & 61.1\% \\
 & & \scriptsize(+$6.7\%\uparrow$) & \scriptsize(+$6.6\%\uparrow$) & \textbf{\scriptsize(+$22.3\%\uparrow$)} & \scriptsize(+$9.8\%\uparrow$) & \scriptsize(+$5.6\%\uparrow$) & \scriptsize(+$10.2\%\uparrow$) \\
 \cmidrule{2-8}
 & \multirow{2}{*}{\textbf{Ours}} & \textbf{73.3\%} & \textbf{63.3\%} & 74.0\% & \textbf{76.7\%} & \textbf{48.0\%} & \textbf{67.1\%} \\
 & & \textbf{\scriptsize(+$23.3\%\uparrow$)} & \textbf{\scriptsize(+$16.6\%\uparrow$)} & \scriptsize(+$21.0\%\uparrow$) & \textbf{\scriptsize(+$11.9\%\uparrow$)} & \textbf{\scriptsize(+$8.1\%\uparrow$)} & \textbf{\scriptsize(+$16.2\%\uparrow$)} \\
\midrule
\multirow{7}{*}{\makecell[l]{\textbf{Qwen3}\\\textbf{-14B}}}
 & Single & 63.3\% & 56.7\% & 64.7\% & 68.5\% & 48.5\% & 60.3\% \\
 \cmidrule{2-8}
 & \multirow{2}{*}{TextMAS} & 63.3\% & 60.0\% & 80.3\% & 72.8\% & 51.5\% & 65.6\% \\
 & & \scriptsize(+0.0\%) & \scriptsize(+$3.3\%\uparrow$) & \scriptsize(+$15.6\%\uparrow$) & \scriptsize(+$4.3\%\uparrow$) & \scriptsize(+$3.0\%\uparrow$) & \scriptsize(+$5.2\%\uparrow$) \\
 \cmidrule{2-8}
 & \multirow{2}{*}{LatentMAS} & 66.7\% & 63.3\% & \textbf{80.7\%} & 75.7\% & 52.0\% & 67.7\% \\
 & & \scriptsize(+$3.4\%\uparrow$) & \scriptsize(+$6.6\%\uparrow$) & \textbf{\scriptsize(+$16.0\%\uparrow$)} & \scriptsize(+$7.2\%\uparrow$) & \scriptsize(+$3.5\%\uparrow$) & \scriptsize(+$7.3\%\uparrow$) \\
 \cmidrule{2-8}
 & \multirow{2}{*}{\textbf{Ours}} & \textbf{70.0\%} & \textbf{66.7\%} & 80.5\% & \textbf{83.3\%} & \textbf{56.5\%} & \textbf{71.4\%} \\
 & & \textbf{\scriptsize(+$6.7\%\uparrow$)} & \textbf{\scriptsize(+$10.0\%\uparrow$)} & \scriptsize(+$15.8\%\uparrow$) & \textbf{\scriptsize(+$14.8\%\uparrow$)} & \textbf{\scriptsize(+$8.0\%\uparrow$)} & \textbf{\scriptsize(+$11.1\%\uparrow$)} \\
\midrule
\multirow{7}{*}{\makecell[l]{\textbf{DeepSeek}\\\textbf{-R1}\\\textbf{-Distill}\\\textbf{-Qwen-7B}}} 
 & Single & 46.7\% & 30.0\% & 38.3\% & 61.6\% & 27.7\% & 40.9\% \\
 \cmidrule{2-8}
 & \multirow{2}{*}{TextMAS} & 46.7\% & 33.3\% & 36.7\% & \textbf{64.3}\% & 35.3\% & 43.3\% \\
 & & \scriptsize(+0.0\%) & \scriptsize(+$3.3\%\uparrow$) & \scriptsize($-1.6\%\downarrow$) & \scriptsize(+$2.7\%\uparrow$) & \scriptsize(+$7.6\%\uparrow$) & \scriptsize(+$2.4\%\uparrow$) \\
 \cmidrule{2-8}
 & \multirow{2}{*}{LatentMAS} & 30.0\% & 16.7\% & 37.8\% & 60.3\% & 29.3\% & 34.8\% \\
 & & \scriptsize($-16.7\%\downarrow$) & \scriptsize($-13.3\%\downarrow$) & \scriptsize($-0.5\%\downarrow$) & \scriptsize($-1.3\%\downarrow$) & \scriptsize(+$1.6\%\uparrow$) & \scriptsize($-6.1\%\downarrow$) \\
 \cmidrule{2-8}
 & \multirow{2}{*}{\textbf{Ours}} & \textbf{56.7\%} & \textbf{40.0\%} & \textbf{40.5\%} & 63.2\% & \textbf{37.9\%} & \textbf{47.7\%} \\
 & & \textbf{\scriptsize(+$10.0\%\uparrow$)} & \textbf{\scriptsize(+$10.0\%\uparrow$)} & \textbf{\scriptsize(+$2.2\%\uparrow$)} & \textbf{\scriptsize(+$1.6\%\uparrow$)} & \textbf{\scriptsize(+$10.2\%\uparrow$)} & \textbf{\scriptsize(+$6.8\%\uparrow$)} \\
\bottomrule
\end{tabular}
\end{table*}

Table~\ref{tab:main_accuracy_percentage} summarizes the main evaluation results.
We compare ReLAT with Single, TextMAS, and LatentMAS across mathematical reasoning, code generation, and knowledge-based QA. Overall, ReLAT achieves the best average accuracy across all evaluated backbone models and improves performance on most individual benchmarks. These results suggest that reconstruction-guided test-time training provides a useful signal beyond standard inference, explicit text-mediated communication, and open-loop latent-state communication.

On mathematical reasoning, ReLAT brings clear gains on the competition-level AIME 2024/2025 benchmarks. For example, with Qwen3-8B, ReLAT improves AIME 2024 accuracy from 56.7\% under LatentMAS to 73.3\%, and improves AIME 2025 accuracy from 53.3\% to 63.3\%. These gains are consistent with our motivation that open-loop latent states may lose query-specific constraints in difficult multi-step reasoning problems, while the reconstruction-guided objective provides an input-anchored training signal.

On code generation, ReLAT also improves performance on MBPP+. At the Qwen3-4B scale, ReLAT achieves 79.9\%, outperforming the same-scale LatentMAS baseline at 73.5\%. At the Qwen3-14B scale, ReLAT further improves MBPP+ accuracy from 75.7\% under LatentMAS to 83.3\%. This result indicates that test-time reconstruction can provide useful instance-specific training  for tasks requiring precise functional alignment.

On knowledge-based QA, ReLAT shows clear gains on GPQA-Diamond and comparable results to LatentMAS on MedQA. On GPQA-Diamond, ReLAT achieves the best result across all Qwen3 scales, reaching 56.5\% with Qwen3-14B. On MedQA, ReLAT performs comparably to LatentMAS across model scales, while both methods improve substantially over Single. 

Beyond the Qwen3 family, ReLAT also improves performance on DeepSeek-R1-Distill-Qwen-7B, achieving the best average accuracy among all four methods. This suggests that reconstruction-guided test-time training generalizes across backbone families with different reasoning characteristics, including reasoning-distilled models. 

Together, these results show that adding a reconstruction-guided closed loop to latent reasoning improves average performance across diverse tasks and backbone settings. Rather than simply transmitting latent states as reliable once produced, ReLAT provides an instance-level training  signal that anchors latent reasoning to the original problem before final answer generation.

\subsection{Comparison with Test-Time Training Method}

To examine whether ReLAT's gains come from the general TTT paradigm or from the proposed reconstruction-guided latent fidelity signal, we compare $\mathcal{L}_{\text{ReLAT}}$ with several alternative test-time optimization objectives. We first compare against two representative TTT baselines: (1) \textbf{COME}~\cite{zhang2024come}, an entropy-based test-time training objective that conservatively minimizes prediction uncertainty; and (2) \textbf{QTTT}~\cite{bansal2025let}, a query-level test-time training strategy that updates model states using local contextual information. We further include a compute-matched \textbf{Direct Reconstruction TTT} baseline, which uses the same temporary LoRA update budget to reconstruct the current test question, but does not construct the differentiable Question $\rightarrow$ Latent Thought $\rightarrow$ Question cycle. Additional implementation details are provided in Appendix~\ref{app:direct_ttt}.

\begin{table*}[t]
\centering
\footnotesize
\setlength{\tabcolsep}{4.5pt}
\renewcommand{\arraystretch}{0.95}

\begin{minipage}[t]{0.45\textwidth}
\vspace{0pt}
\centering
\textbf{(a) Alternative TTT objectives.}
\vspace{3pt}

\resizebox{\linewidth}{!}{
\begin{tabular}{lccc}
    \toprule
    \textbf{Method} & \textbf{AIME25} & \textbf{MBPP+} & \textbf{GPQA-D.} \\
    \midrule
    No TTT & 46.7\% & 64.8\% & 39.9\% \\
    COME & 53.3\% & 75.0\% & 45.5\% \\
    QTTT & 50.0\% & 76.2\% & 47.0\% \\
    \textbf{ReLAT} & \textbf{63.3\%} & \textbf{76.7\%} & \textbf{48.0\%} \\
    \bottomrule
\end{tabular}
}
\end{minipage}
\hfill
\begin{minipage}[t]{0.5\textwidth}
\vspace{0pt}
\centering
\textbf{(b) Direct reconstruction TTT.}
\vspace{3pt}

\resizebox{\linewidth}{!}{
\begin{tabular}{lccc}
    \toprule
    \textbf{Method} & \textbf{AIME25} & \textbf{MedQA} & \textbf{GPQA-D.} \\
    \midrule
    Qwen3-4B+TTT  & 43.3\% & 56.7\% & 31.8\% \\
    Qwen3-4B+\textbf{Ours} & \textbf{56.7\%} & \textbf{67.8\%} & \textbf{42.4\%} \\
    \midrule
    Qwen3-8B+TTT  & 50.0\% & 67.2\% & 40.4\% \\
    Qwen3-8B+\textbf{Ours} & \textbf{63.3\%} & \textbf{74.0\%} & \textbf{48.0\%} \\
    \bottomrule
\end{tabular}
}
\end{minipage}

\caption{
Panel (a) compares ReLAT with alternative TTT objectives under the same backbone setting.
Panel (b) compares ReLAT with a compute-matched direct reconstruction TTT baseline that reconstructs the question without routing the reconstruction loss through the latent thought sequence.
}
\label{tab:ablation_ttt_full}
\end{table*}

As shown in Table~\ref{tab:ablation_ttt_full}, ReLAT outperforms COME and QTTT on average, suggesting that the choice of TTT objective matters. Entropy-based or context-based objectives can improve over the No-TTT baseline, but they do not directly check whether the intermediate latent state preserves the original problem constraints. ReLAT instead uses reconstruction as an input-anchored fidelity signal, encouraging the latent thought sequence $E_\phi(Q)$ to retain the query-specific constraints needed to recover the original query.

The comparison with Direct Reconstruction TTT further shows that ReLAT's gains are not fully explained by question reconstruction alone. Direct Reconstruction TTT can improve performance, but it does not place the reconstruction constraint on the latent thought sequence itself. In contrast, ReLAT routes reconstruction through $E_\phi(Q)$, making the latent state the object being checked and refined during test-time training. This does not guarantee correctness, but it provides a necessary structural check before final answer generation. These results support our central claim that test-time reconstruction is most effective when used as a closed-loop mechanism for latent-state fidelity, rather than as a generic TTT objective.

\subsection{Efficiency Comparison: Latent Optimization vs. Iterative Refinement}

\begin{wrapfigure}{r}{0.4\textwidth}
    \centering
    \includegraphics[width=\linewidth]{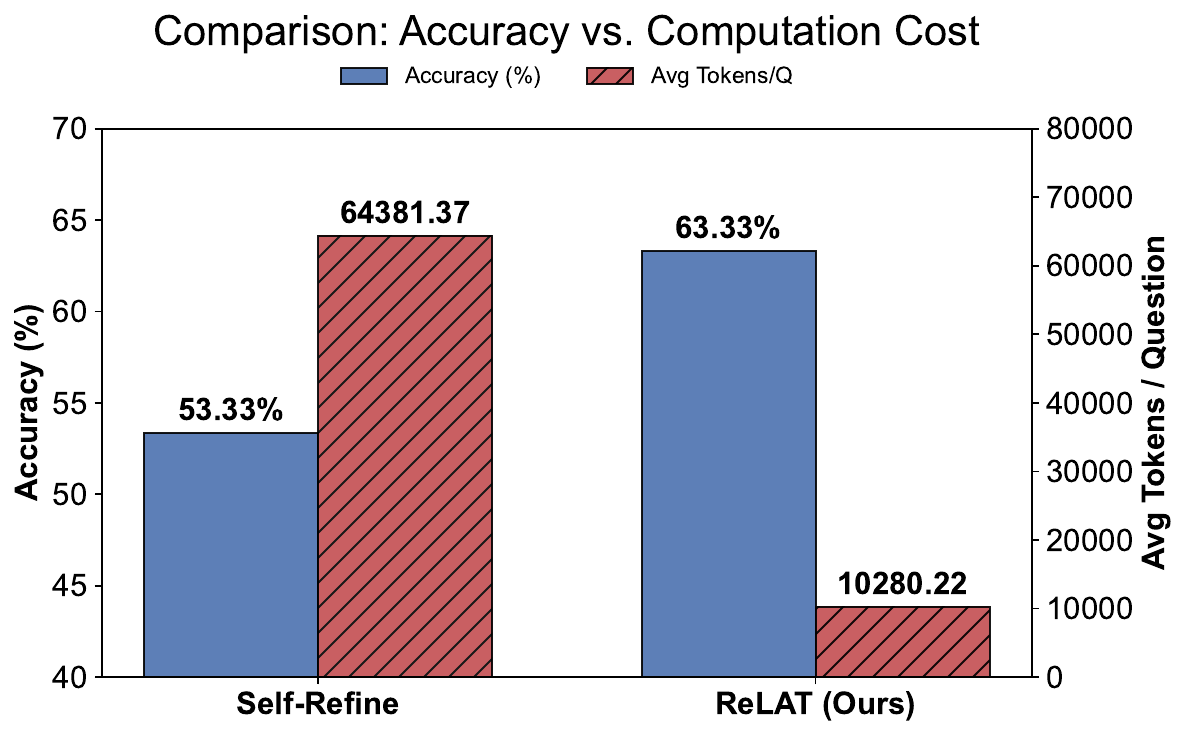}
    \caption{Efficiency on AIME 2025.}
    \label{fig:efficiency_comparison}
\end{wrapfigure}

Text-based iterative reasoning methods often incur substantial token-level overhead because each refinement step requires generating explicit natural-language feedback and revised reasoning traces. To quantify this cost, we compare ReLAT with \textbf{Self-Refine}~\citep{madaan2023self}, a representative iterative verbal refinement baseline. We use Self-Refine as a proxy for text-mediated refinement methods rather than as a strict multi-agent baseline, since its generate-feedback-revise loop captures the main source of token overhead in verbal iterative reasoning.

As illustrated in Figure \ref{fig:efficiency_comparison}, even a simplified comparison with Self-Refine restricted to only three iterations reveals a stark efficiency gap on AIME 2025:

\begin{itemize}
    \item \textbf{Token Explosion}: Self-Refine requires an average of \textbf{64,381.37 tokens} per question to achieve \textbf{53.33\%} accuracy.
    \item \textbf{ReLAT Efficiency}: In contrast, ReLAT achieves a significantly higher accuracy of \textbf{63.33\%} using only \textbf{10,280.22 tokens}.
\end{itemize}

This corresponds to an \textbf{84.0\% reduction in token usage} under our token-counting protocol, while improving accuracy by \textbf{10 percentage points}. These results suggest that ReLAT reduces the token-level communication overhead associated with iterative verbal refinement. Unlike text-based refinement methods, which repeatedly externalize intermediate reasoning in natural language, ReLAT uses a reconstruction-guided test-time training signal to anchor the model to the original problem constraints before final answer generation.

\subsection{Ablation study}

To further analyze the characteristics of our framework, we conducted a series of ablation experiments on the representative Qwen3-8B model. These experiments aim to identify how different hyperparameter configurations—including latent thought length and learning rate—affect the final reasoning accuracy on competition-level math benchmarks.

\begin{table}[ht]
    \centering
    \small
    \setlength{\tabcolsep}{6pt}
    \renewcommand{\arraystretch}{1.08}

    \begin{minipage}[t]{0.43\textwidth}
        \centering
        \textbf{(a) Learning rate}
        \vspace{2pt}

        \begin{tabular}{lcc}
            \toprule
            $\eta$ & AIME 24 & AIME 25 \\
            \midrule
            1e-4 & 53.30\% & 53.33\% \\
            5e-5 & 60.00\% & 63.33\% \\
            2e-5 & 73.33\% & 63.33\% \\
            1e-5 & 63.33\% & 60.00\% \\
            \bottomrule
        \end{tabular}
    \end{minipage}
    \hspace{0.04\textwidth}
    \begin{minipage}[t]{0.43\textwidth}
        \centering
        \textbf{(b) Thought token length}
        \vspace{2pt}

        \begin{tabular}{lcc}
            \toprule
            $K$ & AIME 24 & AIME 25 \\
            \midrule
            4  & 66.67\% & 56.67\% \\
            8  & 63.33\% & 63.33\% \\
            16 & 73.33\% & 63.33\% \\
            32 & 73.33\% & 63.33\% \\
            \bottomrule
        \end{tabular}
    \end{minipage}
    \vspace{5pt}
    \caption{Hyperparameter ablations on AIME 2024 and AIME 2025. $\eta$ denotes the test-time adaptation learning rate, and $K$ denotes the number of latent thought tokens used during latent reasoning.}
    \label{tab:hyperparameter_ablations}
\end{table}

\textbf{Impact of Learning Rate.} We first evaluate the sensitivity of the test-time training process to the learning rate $\eta$. As shown in Table~\ref{tab:hyperparameter_ablations}, the model's performance on AIME 2024 and AIME 2025 varies significantly with the step size. A relatively high learning rate ($1e^{-4}$) causes the performance to drop to 53.30\% on AIME 2024, suggesting that overly aggressive updates can distort the model's internal representations during the short inference window. Conversely, while a very small learning rate ($1e^{-5}$) maintains stability, the improvement is limited due to insufficient adaptation within the given steps. The results indicate that $\eta = 2e^{-5}$ provides the most effective balance for anchoring latent thoughts to the problem constraints.

\textbf{Impact of Thought Token Length.} The capacity of the latent bottleneck, defined by the number of thought tokens $K$, is crucial for balancing information compression and reasoning density. Table~\ref{tab:hyperparameter_ablations} compares the results for $K \in \{4, 8, 16, 32\}$. We observe that a bottleneck of $K=4$ is too restrictive to capture all necessary constraints of complex problems, leading to suboptimal reconstruction. However, increasing the length to $K=32$ does not further improve performance, suggesting that $K=16$ already provides sufficient capacity to preserve the task-relevant constraints for these benchmarks. Considering the additional computational cost introduced by longer latent sequences, we use $K=16$ as the default thought length.

%% file: Sections/Conclusion.tex
We propose \textbf{ReLAT} (\textbf{Re}construction-Guided \textbf{L}atent Reasoning \textbf{A}t \textbf{T}est Time), a self-supervised test-time training framework for improving latent-state fidelity in latent reasoning. ReLAT addresses the open-loop nature of existing latent reasoning methods by constructing a differentiable Question $\rightarrow$ Latent Thought $\rightarrow$ Question reconstruction loop. By optimizing this reconstruction objective at test time, ReLAT provides an input-anchored signal that encourages latent thoughts to preserve the task-specific constraints of the original query before final answer generation.

Across mathematical reasoning, code generation, and knowledge-intensive question answering benchmarks, ReLAT consistently improves over single-model inference, text-based communication, open-loop latent-state communication, and alternative test-time training objectives. These results suggest that reconstruction-guided test-time training is an effective mechanism for closing the loop on opaque latent reasoning states. While reconstruction is not sufficient to guarantee correct reasoning, it provides a useful structural fidelity signal for making latent reasoning more reliable, scalable, and grounded in the original problem specification.

%% file: Sections/Appendix.tex
\section{Prompt Templates and Task Formatting}
\label{app:prompts}

In this section, we present the task instruction templates used in our experiments. These prompts are used for final answer generation across all methods. For ReLAT, the reconstruction objective is applied to the original question tokens while excluding template tokens and formatting artifacts.

\begin{tcolorbox}[colback=white, colframe=blue!70, title=\textbf{Instruction: Mathematics (AIME)}, breakable, arc=1.2mm]
Please reason step by step, and put your final answer within \texttt{\textbackslash boxed\{\}}.
\end{tcolorbox}

\begin{tcolorbox}[colback=white, colframe=blue!70, title=\textbf{Instruction: Multiple-Choice Questions (GPQA / MedQA)}, breakable, arc=1.2mm]
Reason through your answer step-by-step. Then, based on your reasoning, provide the single most likely answer choice. Answer in the format: ``The correct answer is (insert answer here).''
\end{tcolorbox}

\begin{tcolorbox}[colback=white, colframe=blue!70, title=\textbf{Instruction: Code Completion ( MBPP+)}, breakable, arc=1.2mm]
Please complete the function above. Wrap your code in \texttt{```python ... ```}.
\end{tcolorbox}

\section{Case Study: Reasoning Trace Comparison}
\label{app:case_study}

To illustrate how ReLAT can encourage latent states to preserve query-specific constraints, we present a qualitative comparison on a trigonometry phase-shift problem.

\begin{tcolorbox}[colback=red!5, colframe=red!60, title=\textbf{Case 1: Vanilla Qwen3-8B (Open-Loop)}, breakable, arc=1.2mm]
\textbf{Problem:} Find the smallest positive constant $c$ for $y = a \sin(bx + c) + d$ given the graph...

\textbf{Reasoning Trace:}
\begin{enumerate}
    \item Read parameters from code: $a=2, b=3, c=\pi, d=1$.
    \item Wait—the problem asks for the \textit{smallest} $c$. Let me explore periodicity.
    \item Sine is periodic: $\sin(\theta) = \sin(\theta + 2\pi k)$.
    \item Let $c = \pi + 2\pi k$. If $k=-1$, $c = -\pi$ (Not positive).
    \item \textit{[Drift Begins]} Maybe I can use $\sin(\theta) = \sin(\pi - \theta)$? 
    \item Let's check $3x + c = \pi - (3x + \pi)$. This gives $6x + c = 0$. 
    \item This doesn't work for all $x$. Let me re-calculate the phase shift $-c/b$...
    \item \textit{[Redundant Loop]} Let me re-verify $\sin(3x - \pi) = -\sin(3x)$. 
    \item $\sin(3x-\pi) = \sin(3x)\cos(\pi) - \cos(3x)\sin(\pi) = -\sin(3x)$... 
    \item \textit{[Result]} After multiple redundant checks, $c = \pi$.
\end{enumerate}
\end{tcolorbox}

\begin{tcolorbox}[colback=green!5, colframe=green!60, title=\textbf{Case 2: Qwen3-8B with ReLAT (Ours)}, breakable, arc=1.2mm]
\textbf{Reasoning Trace (Optimized by Reconstruction-Guided Test-Time Training):}
\begin{enumerate}
    \item Parameters identified: $a=2, b=3, c=\pi, d=1$.
    \item Goal: Smallest positive $c$ such that $y$ remains identical.
    \item \textbf{[ReLAT Anchor]} The reconstruction loop $Q \to H \to \hat{Q}$ penalizes any $H$ that deviates from the constraint $c > 0$ and graph identity.
    \item Directly check periodicity: $c \in \{ \pi + 2\pi k \mid k \in \mathbb{Z} \}$.
    \item Evaluate boundary: $k=0 \Rightarrow c=\pi$; $k=-1 \Rightarrow c=-\pi$ (Violates $c>0$).
    \item \textbf{[Early Exit]} Since $c = \pi$ is the first positive value in the sequence and the gradient flow from $\mathcal{L}_{cycle}$ confirms no loss of information, the search terminates.
    \item \textbf{Result:} $\boxed{\pi}$. 
\end{enumerate}
\end{tcolorbox}

\section{Direct Reconstruction TTT Baseline}
\label{app:direct_ttt}
The Direct Reconstruction TTT baseline uses the same backbone, LoRA configuration, optimizer, learning rate range, and number of test-time update steps as ReLAT. The difference is that the baseline reconstructs the current query directly without routing the reconstruction loss through a latent thought sequence. Thus, it tests whether ReLAT's gains are explained by generic input reconstruction during test-time training.

Formally, Direct Reconstruction TTT optimizes:
\[
\mathcal{L}_{\mathrm{direct}} = - \sum_{k \in \mathcal{M}(Q)}
\log P(q_k \mid q_{<k}; \theta, \phi),
\]
whereas ReLAT optimizes:
\[
\mathcal{L}_{\mathrm{ReLAT}} = - \sum_{k \in \mathcal{M}(Q)}
\log P(q_k \mid \mathcal{E}_{\phi}(Q), q_{<k}; \theta, \phi).
\]
The comparison isolates whether reconstruction is more effective when imposed as a fidelity constraint on the latent thought sequence.

\section{Runtime and Memory}
\label{app:GPU}
To complement the token-level cost comparison in Section~5.4, we report wall-clock time and peak GPU memory on AIME 2024 using Qwen3-8B on NVIDIA L40S GPUs. The reported runtime includes the full per-example inference pipeline, including test-time optimization when applicable and final answer generation.

\begin{table}[t]
    \centering
    \small
    \setlength{\tabcolsep}{6pt}
    \begin{tabular}{lccc}
        \toprule
        \textbf{Method} & \textbf{Wall-clock / sample} & \textbf{Peak memory} & \textbf{Memory overhead} \\
        \midrule
        Vanilla & 434.34 s & 17,396 MB & -- \\
        Direct Reconstruction TTT & 836.69 s & 17,898 MB & +2.9\% \\
        ReLAT & 1167.55 s & 18,624 MB & +7.1\% \\
        \bottomrule
    \end{tabular}
    \vspace{4pt}
    \caption{Wall-clock time and peak GPU memory on AIME 2024 with Qwen3-8B using NVIDIA L40S GPUs.}
    \label{tab:runtime_memory}
\end{table}

ReLAT introduces additional wall-clock cost because it performs gradient-based adaptation at test time. However, its peak memory overhead remains moderate compared with vanilla inference, since the backbone model is frozen and only temporary LoRA parameters are updated.

\section{Limitations}
\label{app:limitations}
ReLAT is designed to provide an input-anchored fidelity signal for latent reasoning rather than a complete verifier of final answers. Future work could combine reconstruction-guided adaptation with answer verification, tool-based checking, or stronger semantic constraint extraction.

Our experiments mainly focus on several open-source backbone models and representative reasoning, QA, and code benchmarks. Extending ReLAT to larger models, multilingual settings, long-context tasks, and interactive agent environments would further clarify its generality.

%\section{Broader Impacts}
%\label{app:broderimpacts}

%ReLAT may help improve the reliability of latent reasoning systems by reducing semantic drift between an input query and its latent intermediate state. This could benefit reasoning, question answering, and code generation applications where preserving task-specific constraints is important. However, because the method improves model reasoning capability, it could also be used in systems that generate incorrect or harmful outputs more effectively if deployed without appropriate verification. ReLAT should therefore be combined with answer checking, human oversight, and domain-specific safeguards in consequential settings.

%\section{Use of AI Assistants}
%\label{app:llm}

%AI assistants were used as auxiliary tools for manuscript preparation, including language polishing, clarity improvement, and organization. All experimental design, methodological decisions, analyses, reported results, and final content were reviewed and verified by the authors.